\documentclass[journal, compsoc]{IEEEtran}
\usepackage{graphicx,array}
\usepackage{amsmath,amssymb,amsfonts}
\usepackage{multirow}
\usepackage{caption}
\usepackage{xcolor}
\usepackage[table]{xcolor}
\usepackage[inline]{enumitem}

\usepackage{titlesec}

\titlespacing*{\section}{0pt}{0.8\baselineskip}{0.6\baselineskip}
\titlespacing*{\subsection}{0pt}{0.8\baselineskip}{0.4\baselineskip}

\title{MicroViTv2: Beyond the FLOPS for Edge Energy-Friendly Vision Transformers}
\usepackage{hyperref}
\hypersetup{
    colorlinks=true,
    linkcolor=blue,
    filecolor=magenta,      
    urlcolor=cyan,
    pdftitle={Overleaf Example},
    pdfpagemode=FullScreen,
    }
    
\author{Novendra Setyawan, \IEEEmembership{Student Member, IEEE}, Chi-Chia Sun, \IEEEmembership{Member, IEEE}, \\ Mao-Hsiu Hsu, \IEEEmembership{Member, IEEE}, Wen-Kai Kuo,  \IEEEmembership{Member, IEEE}, Jun-Wei Hsieh, \IEEEmembership{Senior Member, IEEE} \\
\IEEEauthorblockA{} 
\thanks{Novendra Setyawan, Mao-Hsiu Hsu and Wen-Kai Kuo are with Department of Electro-Optics, National Formosa University, Taiwan;

Novendra Setyawan also with Department of Electrical Engineering University of Muhammadiyah Malang, Indonesia; 

Chi-Chia Sun is with Department of Electrical Engineering, National Taipei University, Taiwan; 

Jun-Wei Hsieh is with College of Artificial Intelligence and Green Energy, National Yang Ming Chiao Tung University, Taiwan; 

Corresponding Author is Chi-Chia Sun (\textit{E-mail: chichiasun@gm.ntpu.edu.tw}); \\
This work was supported by the National Science and Technology Council, Taiwan under Grant NSTC-113-2221-E-305-018-MY3.}}

\begin{document}
\maketitle
\markboth{Accepted and To be Appear at 2026 IEEE International Conference on Image Processing (ICIP), 13-17 Sept. 2026, Tampere, Finland}%
{Setyawan \etal{}: FaceLiVTv2}

\begin{abstract}

The Vision Transformer (ViT) achieves remarkable accuracy across visual tasks but remains computationally expensive for edge deployment. This paper presents MicroViTv2, a lightweight Vision Transformer optimized for real-device efficiency. Built upon the original MicroViT, the proposed model is designed based on reparameterized design, specifically Reparameterized Patch Embedding (RepEmbed) and Reparameterized Depth-Wise convolution mixer (RepDW) for faster inference, and introduces the Single Depth-Wise Transposed Attention (SDTA) to capture long-range dependencies with minimal redundancy. Despite slightly higher FLOPs, MicroViTv2 improves accuracy up to 0.5\% compared to its predecessor and surpassing MobileViTv2, EdgeNeXt, and EfficientViT while maintaining fast inference and high energy efficiency on Jetson AGX Orin. Experiments on ImageNet-1K and COCO demonstrate that hardware-aware design and structural re-parameterization are key to achieving high accuracy and low energy consumption, validating the need to evaluate efficiency beyond FLOPs. Code is available at \href{https://github.com/novendrastywn/MicroViT}{\textit{https://github.com/novendrastywn/MicroViT}}.

\end{abstract}

\section{Introduction}
In recent years, ViT~\cite{dosovitskiy2020image} has become a powerful paradigm in computer vision, achieving state-of-the-art results across tasks such as image classification, face recognition, detection, and segmentation~\cite{Setyawan2025FaceLiVT, setyawan2026facelivtv2, wang2024smiletrack, hsu2024inpainting}. However, their complexity and large parameter counts make them unsuitable for resource-constrained platforms such as mobile and edge devices. Deploying ViT in these environments requires models that balance accuracy, efficiency, and latency while operating under strict energy budgets.
\begin{figure}[t!]
    \centering
    \includegraphics[width=\columnwidth]{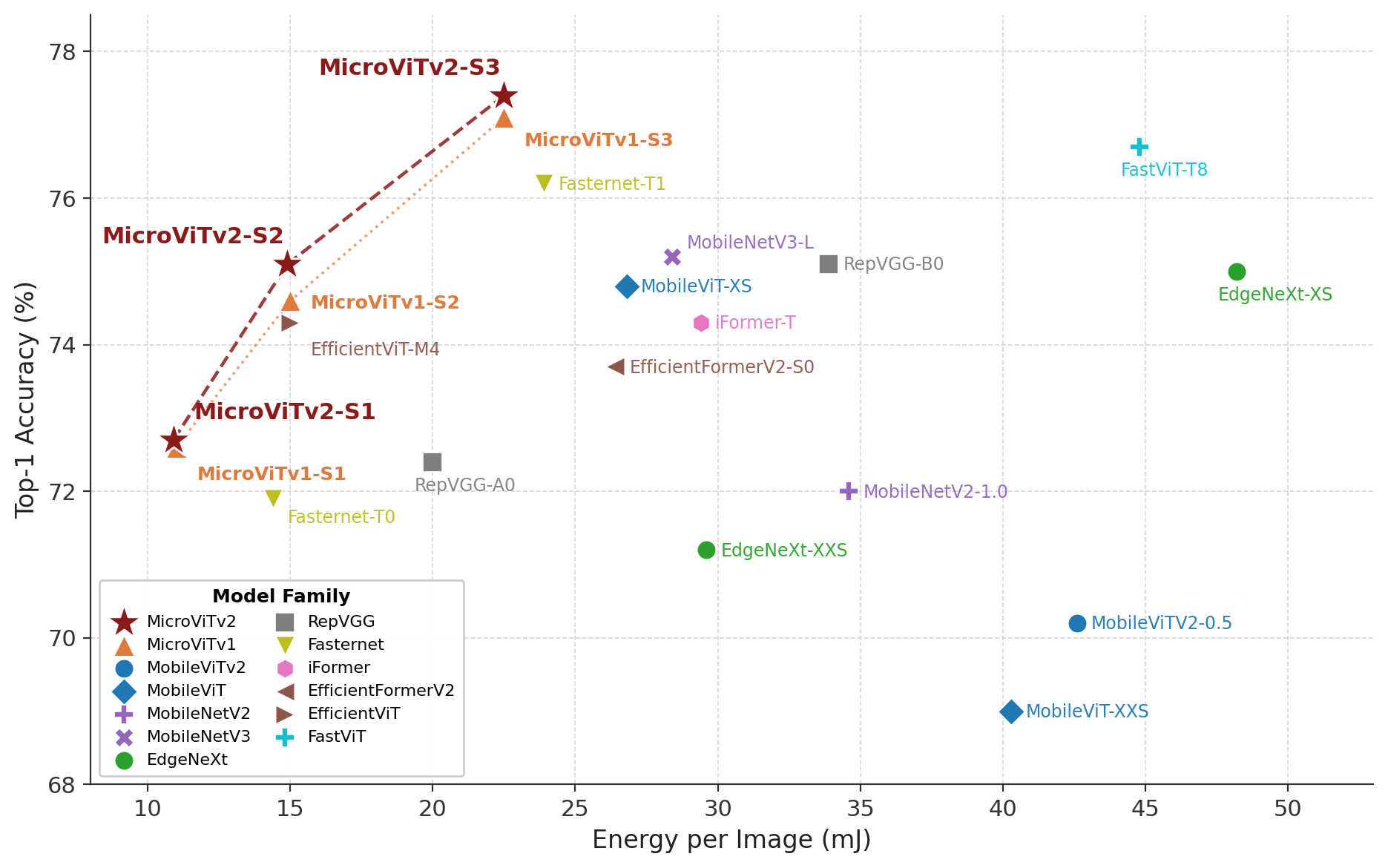}
    \caption{Comparison of our proposed MicroViTv2 model with SOTA methods in accuracy v.s. Edge energy. In theory, the optimal performance is in the upper-left of the plot, which means higher accuracy and less energy usage.}
    \label{fig:plot_pareto}
\end{figure}
\begin{figure*}
    \centering
    \includegraphics[width=18cm]{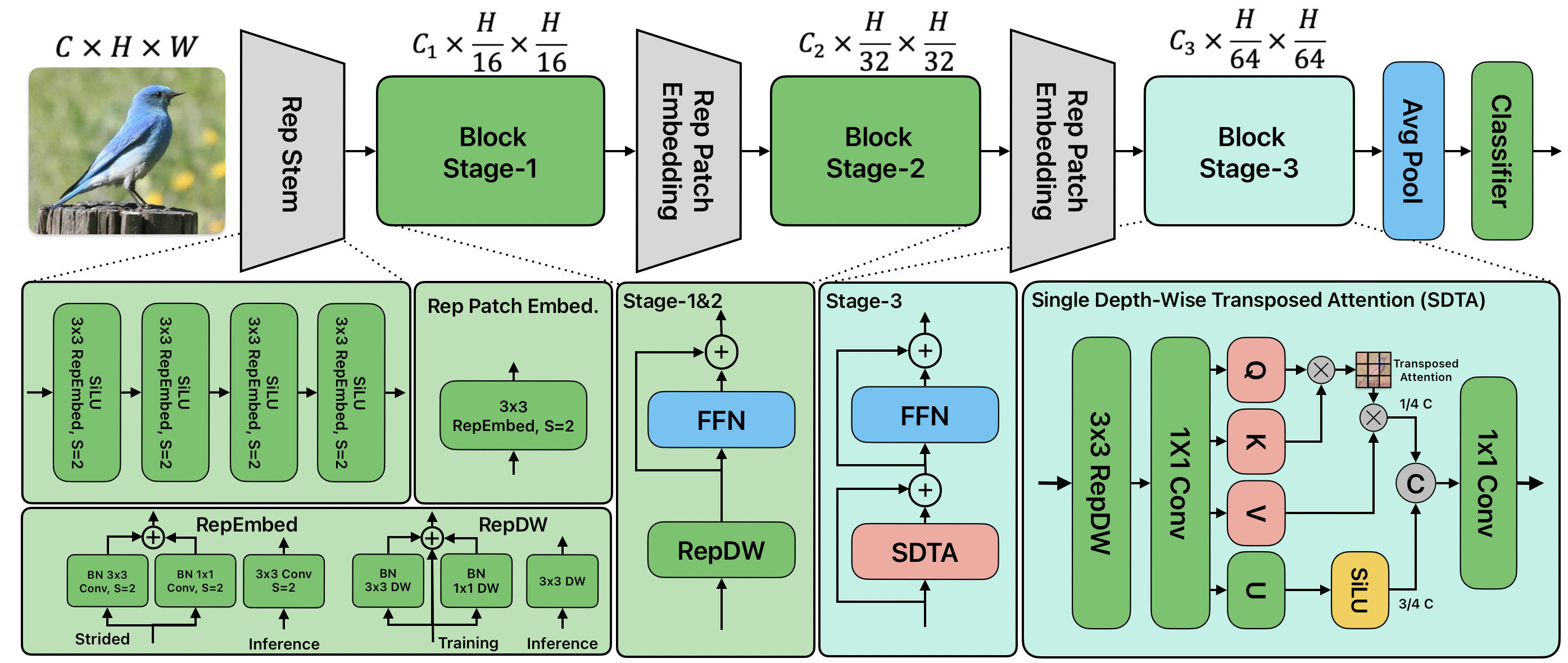}
    \caption{\textbf{MicroViTv2} conducted with 3 stage pyramid feature map architecture. The first two stage use RepDW token mixer and the last stage use Single Dept-Wise Transpose Attention (SDTA).}
    \label{fig:MicroViT-Arch}
\end{figure*}
Several lightweight variants, such as MobileViT \cite{mehta2021mobilevit}, MobileViTv2 \cite{mehta2022separable}, EdgeNeXt \cite{maaz2022edgenext}, FastViT \cite{vasu2023fastvit}, and iFormer \cite{zheng2025iformer} have attempted to bridge this gap by combining convolutional inductive biases with efficient attention modules. Our earlier work  MicroViT \cite{setyawan2025microvit} introduced Efficient Single-Head Attention (ESHA) by leveraging group convolutions and low-redundancy attention. MicroViT, which we now refer to as MicroViTv1, significantly reduced computational cost and power consumption while retaining competitive accuracy. Yet, these approaches often maintain a strong dependency on FLOPs-based metrics when justifying efficiency. This assumption can be misleading: a model with lower FLOPs may still incur higher latency or energy costs on real hardware due to memory access, operator scheduling, or implementation overhead \cite{achmadiah2025energy}. Thus, designing edge-friendly Transformers requires moving beyond FLOPs as the sole efficiency indicator.

Building on these insights, we present MicroViTv2, an enhanced lightweight Transformer that further improves the accuracy-efficiency trade-off through structural reparameterization and attention refinement. Specifically, MicroViTv2 replaces the stem and patch embedding with RepEmbed, token mixers with RepDW, enabling efficient spatial down-sampling and feature extraction during training and streamlined inference via structural re-parameterization. Furthermore, we introduce an SDTA mechanism, which leverages transposed attention to capture long-range dependencies at channel based complexity while dynamically balancing local and global representations. Unlike existing models that equate low FLOPs with efficiency, MicroViTv2 demonstrates that a carefully designed architecture with slightly higher FLOPs can still be faster and more energy-efficient in practice, as shown in Fig. \ref{fig:plot_pareto}.

Comprehensive experiments demonstrate that MicroViTv2 outperforms its predecessor and several  lightweight Transformers on ImageNet-1K classification and COCO detection tasks. On the Jetson AGX Orin, MicroViTv2 achieves improved throughput, lower energy per image, and higher accuracy compared to SOTA, highlighting its effectiveness as an edge-friendly backbone. 
The main contributions of this paper can be summarized as follows:
\begin{enumerate}
    \item We propose MicroViTV2, a structurally re-parameterized lightweight Vision Transformer for edge devices, combining a RepDW-based design with an efficient attention (SDTA) module.
    \item We provide detailed evaluations on edge devices, demonstrating superior trade-offs in throughput, latency, and energy efficiency compared to state-of-the-art lightweight models.
\end{enumerate}
Through these advances, MicroViTv2 represents a step forward in designing practical, energy-aware Vision Transformers for edge computing scenarios.
\section{Related Works}

ViT~\cite{dosovitskiy2020image} has achieved strong performance in computer vision but remains computationally expensive for edge deployment. Several lightweight variants, such as EdgeNeXt~\cite{maaz2022edgenext}, merge cnn with transformers for low-power vision tasks, and SHViT~\cite{yun2024shvit} simplifies multi-head self-attention into a single-head variant to eliminate redundancy. FastViT~\cite{vasu2023fastvit} and EfficientFormerV2~\cite{li2023rethinking} utilize structural re-parameterization and group attention to improve inference efficiency on mobile processors. Our previous work, MicroViTv1~\cite{setyawan2025microvit}, introduced an Efficient Single-Head Attention (ESHA) mechanism that reduced redundancy through grouped convolutions, achieving competitive accuracy and energy savings on edge devices. However, ESHA primarily focused on channel sparsity and offered limited global context modeling. Hence, MicroViTv2 enhances spatial representation and long-range dependency modeling with RepEmbed, RepDW, and SDTA. Reparameterization fuses multi-branch training-time graphs into single inference-time convolutions, improving training convergence while suppressing redundant DRAM transfers at inference;
SDTA performs attention along a fixed channel axis to bound the attention-induced memory footprint. Unlike prior methods that equate efficiency with FLOPs reduction, MicroViTv2 emphasizes real-device performance, achieving higher throughput and lower energy usage, and validating the need to evaluate efficiency beyond FLOPs.

\section{Proposed Method}
The MicroViTv2 model incorporates the reparameterize-based transformer design combined with Single Depth-Wise Transposed Attention (SDTA).

\subsection{Reparameterize Based Design}
As depicted in Fig. \ref{fig:MicroViT-Arch}, MicroViTv1~\cite{setyawan2025microvit} used a sequence of depth-wise and point-wise layers followed by a residual FFN for token embeddings. While efficient in FLOPs, these layers tend to double memory access due to the residual connection. MicroViTv2 develops it with a structurally re-parameterized block. 
\begin{figure}
    \centering
    \includegraphics[width=8cm]{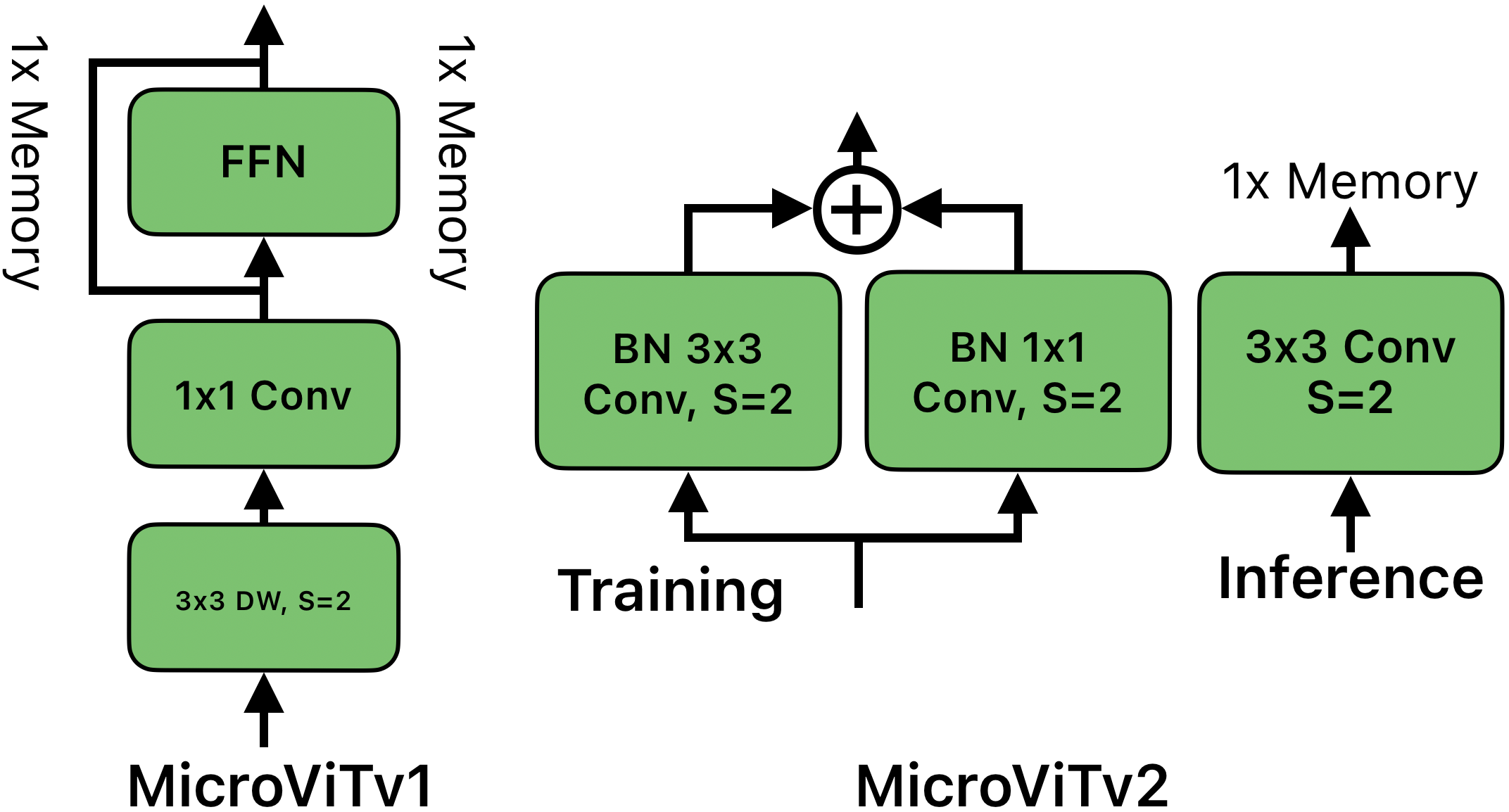}
    \caption{Patch Embedding of MicroViTv1 vs Rep Patch Embedding MicroViTv2}
    \label{fig:MicroViT-Arch}
\end{figure}
First, each patch embedding block (including stems) changes into a reparameterized patch embedding (RepEmbed) that consists of parallel $3\times3$, $1\times1$, and identity branches during training:
\begin{equation}
Y = \mathrm{BN}\big(\mathrm{Conv}_{3\times3}(X) + \mathrm{Conv}_{1\times1}(X) + X\big),
\end{equation}
where $X$ denotes the input feature. Batch normalization (BN) layers are applied after each branch to stabilize optimization. During inference, these branches are mathematically fused into a single equivalent convolution following the reparameterization procedure on ~\cite{Setyawan2025FaceLiVT}. While it adds more convolutional operations during training, its re-parameterization merges these branches into a single convolution at deployment, enhancing both speed and feature extraction efficiency by reducing memory access overhead on GPUs and NPUs.

Following the RepEmbed, the early transformer stages of MicroViT-V2 adopt the reparameterized depth-wise convolution (RepDW) blocks to further enhance spatial representation while preserving energy efficiency. The RepDW design extends the re-parameterization concept of RepEmbed into a depth-wise separable form. It is employed in the first and second pyramid stages to perform efficient spatial mixing before global modeling by SDTA in Stage-3. Each stage follows a residual configuration with a depth-wise spatial mixer and a feed-forward network (FFN):
\begin{align}
    X'  =& X + \mathrm{RepDW}(X),\\
    X'' =& X' + \mathrm{FFN}(X').
\end{align}

 This structure maintains a consistent gradient flow, stable training, and is more suitable for lightweight edge deployment.

\subsection{Single Depth-Wise Transposed Attention (SDTA)}
SDTA is proposed to boost the representational abilities of lightweight Vision Transformers by capturing both local and global dependencies with low computational demands. Evolving from the ESHA in MicroViTv1~\cite{setyawan2025microvit}, which reduced redundancy via grouped convolutions and single-head attention but focused on channel-level sparsity with spatially reduced tokens, limiting it in later stages, SDTA introduces transposed attention paired with depth-wise convolutions. This design allows for improved modeling of global dependencies while maintaining ESHA's efficiency and single-head characteristics.

Let the input feature map of the $i$ -th block be $X_i \in \mathbb{R}^{H \times W \times C}$. The query, key, value, and local-branch tensors are obtained via a sequence of $1\times1$ convolutions with a $\mathrm{RepDW}(\cdot)$ to extract localized spatial features, then split as
\begin{equation}
Q, K, V, U = \mathrm{Split}\big(W_p * RepDw(X_i)\big),
\end{equation}
where $Q, K \in \mathbb{R}^{16\times HW}$, $V \in\mathbb{R}^{\frac{C}{4}\times HW}$, $U \in\mathbb{R}^{\frac{3C}{4}\times H \times W}$ and $W_p$ denote learnable projection weights. The attention map is then computed as
\begin{equation}
Att = V\left(\mathrm{Softmax}\left(\frac{Q^{\mathsf{T}}K}{\sqrt{16}}\right)\right),
\end{equation}
To efficiently merge the local and global information, SDTA concatenates both feature branches and projects them through a $1\times1$ convolution:
\begin{equation}
X'_i = W_o * \mathrm{Cat}\big(Att, \sigma(U)\big),
\end{equation}
where $W_o$ denotes the output projection matrix. This operation has linear complexity by applying the dot-product operations of the attention across channel dimensions (16 as the default of $Q$ and $K$ dimensions)  instead of the spatial dimension. An FFN is then applied after the residual addition:
\begin{align}
    X'  =& X + \mathrm{SDTA}(X),\\
    X'' =& X' + \mathrm{FFN}(X').
\end{align}

Overall, SDTA provides a more expressive and hardware-efficient alternative to ESHA.

\subsubsection{Evaluation Metrics}
Beyond Top-1 accuracy (Acc), we report per-image energy ($E_{\text{img}}$) and the accuracy-energy trade-off ($\eta$) to jointly assess predictive performance and on-device energy cost:
\begin{equation}
E_{\text{img}} = \frac{1000 \cdot \bar{P} \cdot \sum_{i=1}^{N} t_i}{N \cdot B},
\qquad
\eta = \frac{\text{Acc}}{E_{\text{img}}},
\label{eq:energy_eta}
\end{equation}
where $N$ is the number of inference iterations in the measurement window, $B$ the batch size, $t_i$ the latency (s) of the $i$-th iteration, and $\bar{P}$ the mean GPU+SoC power (W) sampled on the Jetson. $E_{\text{img}}$ is reported in mJ and $\eta$ in \%/(mJ/Img).

\begin{table*}[!ht]
\centering 
\setlength{\tabcolsep}{6.0pt} 
\renewcommand{\arraystretch}{1} 
\caption{Benchmarking MicroViTv2's with SOTA on ImageNet-1K, evaluation measured on NVIDIA Jetson AGX Orin}
\begin{tabular}{ lcccccccccc }
\hline
\multirow{2}{*}{Model} & \multirow{2}{*}{Type} & Inp. & Param. & FLOPs & Acc. & Latency & Throughput & Power & Energy  & $\eta$ \\ 
  &  & Res. & (M) & (M) & (\%) & (ms) & (Img/s) & (W) & (mJ/Img) & (\%/mJ/Img) \\ \hline
MobileViTV2-0.5\cite{mehta2022separable}   & Hybrid & 256 & 1.4 & 480 & 70.2 & 5.29 & 666.3  & 28.4 & 42.6  & 1.65 \\
MobileViT-XXS\cite{mehta2021mobilevit}     & Hybrid & 256 & 1.3 & 261 & 69.0 & 4.34 & 699.5  & 28.2 & 40.3  & 1.71 \\
MobileNetV2-1.0\cite{sandler2018mobilenetv2} & CNN  & 224 & 3.5 & 314 & 72.0 & 2.58 & 781.7  & 27.1 & 34.6  & 2.08 \\
EdgeNeXt-XXS\cite{maaz2022edgenext}        & Hybrid & 256 & 1.3 & 261 & 71.2 & 4.65 & 865.2  & 25.6 & 29.6  & 2.41 \\
RepVGG-A0\cite{ding2021repvgg}             & CNN    & 224 & 8.3 & 1.4 G & 72.4 & 3.1 & 1562.9  & 31.9 & 20.0 & 3.62 \\
Fasternet-T0\cite{chen2023run}             & CNN    & 224 & 3.9 & 340 & 71.9 & 4.00 & 1955.2 & 28.1 & 14.4 & 4.99 \\
SHViT-S1 \cite{yun2024shvit}               & Hybrid & 224 & 6.3 & 241 & 72.8 & 6.41 & 2110.9 & 24.2 & 11.5 & 6.33 \\
MicroViTv1-S1 \cite{setyawan2025microvit}    & Hybrid & 224 & 6.4 & 231 & 72.6 & 4.84 & 2366.9 & 26.0 & 11.0 & 6.60 \\ 
\rowcolor{gray!25}  
MicroViTv2-S1                              & Hybrid & 224 & 6.7 & 250 & 72.7 & 4.84 & 2367.6 & 25.8 & 10.8 & 6.67 \\ 
\hline
EdgeNeXt-XS\cite{maaz2022edgenext}         & Hybrid & 256 & 2.3 & 536 & 75.0 & 5.96 & 573.9  & 27.6 & 48.2 & 1.56 \\
RepVGG-B0\cite{ding2021repvgg}             & CNN    & 224 & 14.3 & 3.1 G & 75.1 & 4.20 & 997.1  & 33.9 & 33.9 & 2.21 \\
iFormer-T\cite{liu2023efficientvit}        & Hybrid & 224 & 2.9 & 530 & 74.3 & 4.12 & 1016.7 & 29.8 & 29.4 & 2.53 \\
MobileNetV3-L\cite{howard2019searching}    & CNN    & 224 & 3.5 & 314 & 75.2 & 3.49 & 941.5  & 26.7 & 28.4 & 2.65 \\
EfficientFormerV2-S0\cite{li2023rethinking} & Hybrid & 224 & 3.6 & 407 & 73.7 & 4.64 & 900.7 & 30.8 & 26.4 & 2.79 \\
MobileViT-XS\cite{mehta2021mobilevit}      & Hybrid & 256 & 2.3  & 935 & 74.8 & 4.75 & 376.1 & 30.7 & 26.8  & 2.79 \\
SHViT-S2 \cite{yun2024shvit}               & Hybrid & 224 & 11.5 & 366 & 75.2 & 6.56 & 1666.1 & 25.3 & 15.2 & 4.95 \\
EfficientViT-M4 \cite{liu2023efficientvit} & Hybrid & 224 & 8.8  & 303 & 74.3 & 9.17 & 1687.1 & 25.1 & 15.0 & 4.95 \\
MicroViTv1-S2 \cite{setyawan2025microvit}    & Hybrid & 224 & 10.0 & 345 & 74.6 & 5.25 & 1843.2 & 27.6 & 15.0 & 4.97 \\
\rowcolor{gray!25}
MicroViTv2-S2                             & Hybrid & 224 & 12.7 & 407 & 75.1 & 5.25 & 1883.3 & 28.0 & 14.9 & 5.04 \\ 
\hline
FastViT-T8\cite{vasu2023fastvit}          & Hybrid & 256 & 4.0 & 687 & 76.7 & 2.67 & 665.6 & 29.8 & 44.8  & 1.70 \\
Fasternet-T1\cite{chen2023run}            & CNN    & 224 & 7.6 & 851 & 76.2 & 4.41 & 1314.8 & 31.4 & 23.9 & 3.19 \\
SHViT-S3\cite{yun2024shvit}               & Hybrid & 224 & 14.1 & 366 & 77.4 & 7.35 & 1245.9 & 29.0 & 22.6 & 3.42 \\
MicroViTv1-S3\cite{setyawan2025microvit}    & Hybrid & 224 & 16.7 & 580 & 77.1 & 5.90 & 1326.2 & 29.8 & 22.5 & 3.43 \\
\rowcolor{gray!25}
MicroViTv2-S3  & Hybrid & 224 & 17.0 & 676 & 77.4 & 5.90 & 1335.5 & 29.9 & 22.5 & 3.44 \\ 
\hline 
\end{tabular}
\label{tab:edge-result}
\end{table*}

\begin{table}[!ht]
\centering 
\setlength{\tabcolsep}{4.0pt} 
\caption{All Variant MicroViTv2 Model configurations.}
\begin{tabular}{ lccc }
\hline
Variant & Depth & Dims & FFN exp ratio \\  
 \hline
MicroViTv2-S1  & $[3, 8, 5]$ & $[128, 224, 320]$ & 2 \\
MicroViTv2-S2  & $[3, 9, 5]$ & $[128, 224, 448]$ & 2 \\
MicroViTv2-S3  & $[4, 9, 6]$ & $[128, 384, 448]$ & 2 \\
\hline
\end{tabular}
\label{tab:variant}
\end{table}

\section{Result}
For the evaluation of all variants MicroViTv2 in \ref{tab:variant}, ImageNet-1K dataset \cite{russakovsky2015imagenet} was used, comprising 1.28 million training images and 50,000 validation images in 1,000 categories. Following the DeiT training method \cite{touvron2021training}, the models were trained for 300 epochs at a 224×224 resolution with an initial learning rate of 0.004, utilizing various data augmentations. AdamW Optimizer was used with a batch size of 512 on four 4090 GPUs.

We evaluated the latency, throughput, and energy usage of the model in the Jetson AGX Orin edge device. For throughput, we used a batch size of 64 with ONNX Runtime, while the latency was evaluated in a single batch size. To enhance performance during inference, we fused the BN layers with adjacent layers when possible. We examined power and energy usage during throughput tests with a 60-second duration at a consistent resolution.

We further evaluate MicroViTv2 on object detection on the COCO dataset~\cite{lin2014microsoft} utilizing RetinaNet~\cite{lin2017focal} and conduct training for 12 epochs (1$\times$ schedule), adhering to the configuration used by \cite{liu2023efficientvit} in MMDetection. In the object detection experiments, we employ AdamW with a batch size of 16, a learning rate of $1\times10^{-3}$, and a weight decay rate of 0.025. 

\subsection{ImageNet-1K Classification Result}
Table \ref{tab:edge-result} presents a comparison of various MicroViTv2 variants with state-of-the-art (SOTA) models on the ImageNet-1K dataset. The evaluation focuses on models' computational efficiency and accuracy, highlighting the trade-offs between resource consumption and performance. 

Across all variants, MicroViTv2 consistently achieves higher Top-1 accuracy and throughput than its predecessor and other lightweight baselines. In particular, MicroViTv2-S2 improves accuracy by 0.4\% and throughput by  0.26\% over MicroViTv1-S2, while MicroViTv2-S3 achieves 77.4\% Top-1 with a throughput of 1335.5 images/s, faster than FastViT-T8 despite using fewer parameters.

Although MicroViTv2 variants have slightly higher FLOPs than MicroViT, they deliver significantly better throughput and energy efficiency. For example, MicroViTv2-S2 achieves 1883.3 Img/s and $\eta=5.04$, compared to 1843.2 Img/s and $\eta=4.97$ for MicroViTv1-S2. This contradicts the common assumption that lower FLOPs guarantee higher efficiency. Instead, the improvement stems from RepEmbed, RepDW, and SDTA’s linear attention, which better utilize hardware and reduce memory-access latency.
 
On the Jetson Orin, MicroViTv2 achieves lower energy per image than other hybrid models such as EdgeNext-XS ($3.2\times$) and FastViT ($2\times$). The S1 configuration consumes only 10.9 mJ per image, ranking among the most power-efficient in its class. These results validate the effectiveness of structural optimization over arithmetic simplification in practical energy-aware deployment.
 
In general, MicroViTv2 surpasses both convolutional and transformer-based competitors in the balance of accuracy, throughput, and energy cost. Although FLOPs increase marginally by 7 \%, real-world performance gains are substantially higher, underscoring that \textit{efficiency must be assessed beyond FLOPs}. This finding confirms that memory scheduling, kernel fusion, and structural re-parameterization are more decisive factors for edge-device viability than arithmetic operation counts.

\subsection{Object Detection Result}
To further validate the generalization ability of the proposed backbone, MicroViTv2 was integrated into RetinaNet~\cite{lin2017focal} for object detection on the COCO~\cite{lin2014microsoft} \texttt{val2017} split. Training followed the standard $1\times$ schedule using AdamW optimizer with a learning rate $2e-4$ and an input resolution of $1280\times800$. The learning rate will be reduced with a decay of 0.1 in epochs 8 and 11. An inference test on GPU RTX-4090 was also conducted to show the different approaches of MicroViTV1 and MicroViTV2. Table~\ref{tab:obj} presents the comparison against representative lightweight backbones. Despite using slightly more parameters and FLOPs than MicroViTv1-S3, MicroViTv2-S3 achieves a higher inference speed on GPU with identical $AP^{b}$ performance. This demonstrates that the reparameterization-based design and SDTA modules improve feature utilization without compromising detection accuracy. These results further confirm that energy- and latency-aware architectural design yields superior deployment performance even when theoretical FLOPs increase.

\begin{table}[!t]
\centering 
\setlength{\tabcolsep}{4.0pt} 
\renewcommand{\arraystretch}{1} 
\caption{Object detection on COCO val2017 with RetinaNet. $AP^b$ denote bounding box average precision. The GFLOPs and FPS are measured at resolution 1280 $\times$ 800.
}
\begin{tabular}{ lcccccc}
\hline
Backbone & $AP^{b}$  & $AP^{b}_{50}$ & $AP^{b}_{75}$ & Param & FLOPs & FPS \\ \hline
MobNetV2\cite{sandler2018mobilenetv2} & 28.3& 46.7& 29.3& 3.4& 300G & - \\
MobNetV3\cite{howard2019searching} & 29.9& 49.3& 30.8& 5.4 & 217G & -  \\ 
EffViT-M4\cite{liu2023efficientvit}&32.7 &52.2 &34.1 & 8.8 & 299G & - \\
MicViTv1-S3\cite{setyawan2025microvit} & 36.0 & 56.6 & 38.2 & 26.7 & 159G & 209.1 \\ 
\rowcolor{gray!30}
MicViTv2-S3 & 36.0 & 56.8 & 37.5 & 27.3 & 165G & 214.9    \\ \hline
\end{tabular}
\label{tab:obj}

\end{table}

\subsection{Ablation Study}

Table~\ref{tab:abl-result} reports an ablation of MicroViTv2 on ImageNet-1K using NVIDIA Jetson AGX Orin, isolating each component and additionally substituting SDTA with the Multi-DConv Head Transposed Attention
(MDTA) of Restormer~\cite{zamir2022restormer}. 
The results reveal that an increase in \textit{FLOPs does not necessarily imply higher energy consumption or lower throughput}. For instance, replacing the patch embedding with a $3\times3$ convolution increases FLOPs by 2.6\% from 345M to 354M, yet throughput improves by 5.1\% and energy efficiency $\eta$ increases from 4.97 to 5.33~(\%/mJ/I). Similarly, introducing SDTA increases FLOPs to 368M, mostly due to the changes in group convolution (ESHA), but it achieves both higher accuracy and better throughput, reducing per-image energy from 15.0~mJ to 14.2~mJ. Replacing SDTA with MDTA incurs 7.9\% more FLOPs, with 38\% lower throughput and similar accuracy, resulting in a 43.8\% decrease in accuracy-energy efficiency, confirming the efficient design of SDTA.
 
Across all ablations, FLOPs varied from 345M to 407M, while accuracy improved from 74.6 to 75.1, and energy efficiency $\eta$ increased from 4.97 to 5.04~(\%/mJ/I), showing \emph{no direct correlation between FLOPs and energy efficiency}. These findings confirm that energy-aware model design depends more on \emph{structural design and memory access patterns} than on FLOPs alone. This analysis empirically supports the central hypothesis of MicroViTv2: practical efficiency on edge hardware must be evaluated \textit{beyond FLOPs}. 

\begin{table}[!t]
\centering 
\setlength{\tabcolsep}{2.5pt} 
\renewcommand{\arraystretch}{1} 
\caption{Ablation study of MicroViTv2 on ImageNet-1K Dataset, evaluation measured on NVIDIA Jetson AGX Orin}
\begin{tabular}{ lccccccc }
\hline
\multirow{2}{*}{Ablation} & Par & FLP & Acc  & Thru.   & Pow. & Energy  & $\eta$ \\ 
                          & (M) & (M) & (\%) & (Img/s) & (W)  & (mJ/I)  & (\%/mJ/I) \\ \hline
MicroViTv1-S2 \cite{setyawan2025microvit} & 10.0 & 345 & 74.6 & 1843.2 & 27.6 & 15.0 & 4.97 \\
Patch $\rightarrow$ RepEmbed      & 11.4 & 354 & 74.5 & 1937.2 & 27.1 & 14.0 & 5.31 \\
DWC $\rightarrow$ RepDW           & 11.4 & 354 & 74.6 & 1937.2 & 27.1 & 14.0 & 5.33 \\
ESHA $\rightarrow$ SDTA           & 12.2 & 368 & 74.8 & 1960.5 & 27.9 & 14.2 & 5.28 \\
Depth $\uparrow$                  & 12.7 & 407 & 75.1 & 1883.3 & 28.0 & 14.9 & 5.04 \\
\hline
SDTA $\rightarrow$ MDTA\cite{zamir2022restormer} & 14.7 & 439 & 75.0 & 1168.2 & 30.9 & 26.5 & 2.83 \\
\hline
\end{tabular}
\label{tab:abl-result}
\end{table}

\section{Conclusion}
This work introduced MicroViTv2, a lightweight Vision Transformer optimized for edge devices through the proposed Single Depth-Wise Transposed Attention (SDTA) and reparameterized-based design, especially RepEmbed and RepDW, which  enhance feature extraction and global dependency modeling while maintaining low complexity. Experiments on ImageNet-1K and COCO confirm that MicroViTv2 surpasses prior lightweight models in accuracy, throughput, and energy efficiency on real hardware. Despite higher theoretical FLOPs, it achieves faster inference and lower power usage, proving that FLOPs alone do not represent true efficiency. By integrating hardware-aware design, MicroViTv2 advances the development of energy-efficient, edge-deployable model and establishes a foundation for optimization beyond the FLOPs paradigm. Future work will extend evaluation to general edge hardware and integrate energy modeling for hardware-aware neural architecture search.

\bibliography{ref}
\bibliographystyle{ieeetr}
\end{document}